\catcode`\@=11 
\documentclass[11pt]{article}

\usepackage[margin=1in]{geometry}
\usepackage{comment}
\usepackage{amsmath,amssymb,amsthm,mathtools}

\newtheorem{theorem}{Theorem}[section]

\theoremstyle{definition}
\newtheorem{definition}[theorem]{Definition}

\theoremstyle{remark}

\newcommand{\R}{\mathbb{R}}

\newcommand{\Loss}{\mathcal{L}}
\newcommand{\Wuc}{W^{*_{\text{UC}}}}
\newcommand{\Wprime}{W^{\prime}}

\begin{document}

\title{Unit-Consistent (UC) Adjoint for GSD and Backprop in\\
                  Deep Learning Applications}
\author{Jeffrey Uhlmann\\
Dept.\ of Electrical Engineering and Computer Science\\
University of Missouri - Columbia}
\date{}
\maketitle

\begin{abstract}
Deep neural networks constructed from linear maps and positively homogeneous nonlinearities (e.g., ReLU) possess a fundamental gauge symmetry: the network function is invariant to node-wise diagonal rescalings. However, standard gradient descent is not equivariant to this symmetry, causing optimization trajectories to depend heavily on arbitrary parameterizations. Prior work has proposed rescaling-invariant optimization schemes for positively homogeneous networks (e.g., path-based or path-space updates). Our contribution is complementary: we formulate the invariance requirement at the level of the backward adjoint/optimization geometry, which provides a simple, operator-level recipe that can be applied uniformly across network components and optimizer state. By replacing the Euclidean transpose with a Unit-Consistent (UC) adjoint, we derive UC gauge-consistent steepest descent and backprogation.\\ 
\begin{footnotesize}
~\\
\noindent {\bf Keywords}: {\em UC GSD, UC adjoint. Unit Consistency (UC). Deep learning optimization. Parameterization invariance. 
Positive homogeneity (ReLU networks).} 
\end{footnotesize}
\end{abstract}

\section{Unit-Consistent Inverses to UC Adjoints}

A unit-consistent (UC) generalized matrix inverse, $M^{-\text{\tiny U}}$, was developed in \cite{U2018, U2015} that is {\em consistent}
(alt., covariant or equivariant) with respect to arbitrary nonsingular diagonal scalings $D$ and $E$:
\begin{equation*}
     M^{-\text{\tiny U}} ~=~ E\left(D M E\right)^{-\text{\tiny U}}D.
\end{equation*}
or, equivalently:
\begin{equation*}
     (DME)^{-\text{\tiny U}} ~=~ E^{-1} M^{-\text{\tiny U}} D^{-1}.
\end{equation*}
This is not satisfied by the Moore-Penrose (MP) pseudoinverse, which provides consistency with respect to orthonormal/unitary (e.g., rotation) transformations rather that diagonal scale transformations. The MP pseudoinverse (and least-squares minimization more generally) is widely misapplied in contexts that do not satisfy its consistency properties, and it was noted in \cite{U2018} that this includes gradient descent in machine learning in many applications. In this note, we apply analysis analogous to that used to derive the UC inverse to derive a unit-consistent matrix adjoint applicable to machine learning applications involving implicitly UC activation functions.

\paragraph{Related work.}
Several lines of work recognize that standard Euclidean-gradient updates in deep networks can depend strongly on parameterization, including the rescaling symmetries induced by positive homogeneity. Natural-gradient methods address this in principle by defining steepest descent under a non-Euclidean metric \cite{amari1998natural}, and path-normalized approaches such as Path-SGD target ReLU rescaling invariances more directly through a path-based geometry \cite{neyshabur2015pathsgd}. In practice, architectural normalizations (e.g., BatchNorm, LayerNorm) are widely used as effective stabilizers \cite{ioffe2015bn,ba2016ln}, and more recent work shows that competitive training without normalization can be possible with carefully designed training rules \cite{brock2021nfnets}. The UC framework is complementary to these approaches. Specifically, rather than proposing a particular metric (as in natural-gradient variants), or a specific invariant norm (as in path-based methods), UC formulates the consistency requirement at the level of operator adjoints and coordinate changes, which provides a general recipe for constructing backward/updates that are equivariant to the underlying diagonal rescaling gauge while leaving the forward architecture unchanged.

\section{The Geometry of Scale}

Deep networks built from linear maps and positively homogeneous nonlinearities (e.g.,ReLU) exhibit a massive degree of gauge freedom. For example, consider a standard bias-free multilayer perceptron:
\begin{equation}
    h_{0}=x,~ z_{l}=W_{l}h_{l-1},~ h_{l}=\sigma(z_{l}),
\end{equation}
where $\sigma = \text{ReLU}$ \cite{nair2010relu}. The identity $\sigma(\alpha u)=\alpha\sigma(u)$ for $\alpha > 0$ implies that the network function is unchanged if the hidden units are rescaled by diagonal matrices $S_l > 0$ satisfying \cite{neyshabur2015pathsgd}:
\begin{equation}
    \tilde{W}_{l}=S_{l}W_{l}S_{l-1}^{-1}.
\end{equation}
While the parameterizations $W$ and $\tilde{W}$ are functionally identical, they are {\em geometrically distinct}. Standard optimization algorithms, such as Stochastic Gradient Descent (SGD), implicitly assume global rotation sysmmetry. This means that their estimates of local gradients are computed based on an incorrect assumed geometry \cite{amari1998natural}, which introduces a non-stochastic local bias. It is non-stochastic in the sense that it results from local structural artifacts of a gauge freedom that the SGD does not recognize\footnote{This is analogous to trying to estimate relative distances between objects in an environment while observing through a warped lens.}, which means that deviations due to violated symmetry assumptions will not tend to ``filter out'' over a sequence of local steps. 

The symptoms of the mismatch between the actual symmetry and that implicitly assumed by most SGD implementations are well known, but instead of correcting the disconnect, methods such as Batch Normalization (BN) \cite{ioffe2015bn} or Layer Normalization (LN) \cite{ba2016ln} are applied to scale activations during the forward pass in a way that moderates the symptoms, but not the underlying cause. 

\subsection{The Dimensional Flaw in the Euclidean Transpose}

The geometric inconsistency\footnote{Here “inconsistency” refers to symmetry/metric mismatch: $W^\top$ is the Euclidean adjoint, but Euclidean geometry is not invariant to the diagonal rescalings that leave the network function unchanged.} in standard backpropagation is use of the Euclidean adjoint, $W^{\top}$. To see this, consider a linear map $y=Wx$ in a unit-analysis context. If inputs $x$ have physical units of [Meters] and outputs $y$ have units of [Seconds], then the operator $W$ carries units $[\text{Seconds}/\text{Meters}]$. For the gradient update $W \leftarrow W - \eta \nabla_W \Loss$ to be physically meaningful, the gradient must have the units as $W$. However, standard backpropagation derives the gradient via the transpose as \cite{rumelhart1986backprop}:
\begin{equation}
    \nabla_x \Loss = W^{\top} (\nabla_y \Loss).
\end{equation}
The Euclidean transpose $W^{\top}$ numerically retains the entries of $W$, but it blindly mixes the dimensions. If the input units are rescaled to ($x \mapsto Ex$), and output units to ($y \mapsto Dy$), the transpose transforms as:
\begin{equation}
    \widetilde{W^{\top}} = (DWE^{-1})^{\top} = E^{-1} W^{\top} D,
\end{equation}
which ``mixes'' the arbitrary diagonal scales $D$ and $E$ into the backward signal. The result is a gradient update that essentially subtracts {\em incommmensurate units} from the weights, which makes the assumed optimization landscape incorrectly sensitive to the choice of measurement units (the gauge). 

\subsection{Contribution: The Unit-Consistent Adjoint}

The contribution of this work is to replace the Euclidean transpose with a \textbf{Unit-Consistent (UC) Adjoint}. We define this operator by projecting the weight matrix $W$ into a canonical coordinate system that is invariant to arbitrary row/column scaling. Unlike the standard transpose, the UC adjoint is invariant to diagonal gauge transformations. It ensures that the backpropagated error signal and resulting parameter updates respect the intrinsic geometry of the network function instead of the arbitrary implicit scaling of its parameters.

\section{The Unit-Consistent Adjoint and UC-GSD}

To achieve full diagonal-scale consistency, both the backward signal and the resulting gradient update must be computed in a canonical coordinate system.

\subsection{Canonical Decomposition}

Let $W \in \R^{m \times n}$ be a linear operator. A canonical diagonal decomposition (e.g., via RZ scaling \cite{U2018}) can be defined as:
\begin{equation}
     W ~=~ D \Wprime E,
\end{equation}
where $D \in \R^{m \times m}, E \in \R^{n \times n}$ are positive diagonal matrices, and $\Wprime = \mathcal{C}(W)$ is the canonical representative (``gauge-fixed'' form) of the equivalence class of $W$. The computation can be performed in optimal $O(m,n)$ time for $m\times n$ $W$ \cite{U2015,U2018}. 

\begin{definition}[Unit-Consistent Adjoint]
Given $W = D \Wprime E$, the Unit-Consistent Adjoint is defined as\footnote{For notational convenience, we use $D$ and $E$ here to define positive canonical scalings instead of as arbitrary diagonal matrices as done earlier}:
\begin{equation}
     \Wuc ~:=~ E^{-1} W^{\top} D^{-1}.
\end{equation}
\end{definition}
Using the identity $W^{\top} = E (\Wprime)^{\top} D$, this simplifies to $\Wuc = (\Wprime)^{\top}$. This can be interpreted as: {\em the UC adjoint is the transpose taken in canonical coordinates.}

\subsection{Derivation of the Equivariant Update}

Standard gradient descent performs updates $W$ using the Euclidean gradient $G = \nabla_W \Loss$. To construct a gauge-equivariant update, steepest descent must be performed on the canonical variable $\Wprime$, with the result then mapped back to the original parameter space:
\begin{enumerate}
\item {\bf Gradient in Canonical Coordinates} - 
Using the chain rule on $W = D \Wprime E$:
\begin{equation}
     \nabla_{\Wprime}\Loss ~=~ D (\nabla_W \Loss) E ~=~ D G E.
\end{equation}

\item {\bf Descent in Canonical Coordinates} - 
Perform a gradient step on $\Wprime$:
\begin{equation}
     (\Wprime)^+ ~=~ \Wprime - \eta \nabla_{\Wprime}\Loss.
\end{equation}

\item {\bf Push-forward to Original Coordinates} -
Reconstruct the updated weight matrix using the original scales $D, E$:
\begin{eqnarray}
     W^+ &=& D (\Wprime)^+ E \\
               &=& D (\Wprime - \eta D G E) E \\
               &=& D \Wprime E - \eta D^2 G E^2.
\end{eqnarray}
\end{enumerate}
Substituting $W = D \Wprime E$ then gives the Unit-Consistent Gauge-Equivariant Steepest Descent (UC-GSD) rule:
\begin{equation} 
\label{eq:update_rule}
     \boxed{ W^+ ~=~ W - \eta D^2 G E^2 }
\end{equation}

\subsection{Proof of Equivariance}

It is straightforward to verify that this update preserves gauge equivalence. Let $\tilde{W} = S_{out} W S_{in}^{-1}$. The canonical factors transform as $\tilde{D} = S_{out}D$ and $\tilde{E} = E S_{in}^{-1}$, and the Euclidean gradient transforms as $\tilde{G} = S_{out}^{-1} G S_{in}$. The preconditioned update term $\tilde{U}$ transforms as:
\begin{eqnarray}
     \tilde{U} &:=& \tilde{D}^2 \tilde{G} \tilde{E}^2 \\
                    &=& (S_{out} D)^2 (S_{out}^{-1} G S_{in}) (E S_{in}^{-1})^2 \\
                    &=& S_{out} (D^2 G E^2) S_{in}^{-1} \\
                    &=& S_{out} U S_{in}^{-1}.
\end{eqnarray}
Thus:
\begin{eqnarray}
     \tilde{W}^+ &=& \tilde{W} - \eta \tilde{U} \\
                          &=& S_{out} (W - \eta U) S_{in}^{-1} \\
                          &=& S_{out} W^+ S_{in}^{-1}. 
\end{eqnarray}     
This establishes that the optimization trajectory is identical regardless of the chosen gauge.

\section{Gauge-Equivariant Extensions to Network Operations}

The UC framework is extended here to obtain analogs for common deep learning primitives that 
maintain global consistency.

\subsection{Biases}
Consider $z = Wx + b$. Forward equivariance requires $b$ to scale identically to the output $z$. If $z \to S_{out} z$, then $b \to S_{out} b$. We define the canonical bias $b' = D^{-1}b$ using the same output scale $D$ derived from $W$.
The update rule follows the same derivation:
\begin{equation}
     b^+ ~=~ b - \eta D^2 (\nabla_b \Loss).
\end{equation}

\subsection{Convolutions}

A convolution kernel $K$ of shape $(C_{out}, C_{in}, kH, kW)$ transforms under the gauge via its channel dimensions:
\begin{equation}
     \tilde{K}_{i,j,u,v} ~=~ (S_{out})_{ii} K_{i,j,u,v} (S_{in}^{-1})_{jj}.
\end{equation}
We apply the canonical scaling decomposition to the matricized kernel (flattening spatial dimensions). The resulting update broadcasts the diagonal preconditioners $D^2$ (acting on output channels) and $E^2$ (acting on input channels) across the spatial dimensions of the gradient:
\begin{equation}
     K_{i,j,u,v}^{+} ~=~ K_{i,j,u,v} - \eta (D^2)_{ii} (\nabla_K \Loss)_{i,j,u,v} (E^2)_{jj}.
\end{equation}

\subsection{Residual Connections}

In a residual block $y = x + F(x)$, the addition operation forces the scale of the input and the output to match \cite{he2016resnet}: $S_y = S_x = S_{F(x)}$.
This implies that for any learnable layer within the residual branch $F(x)$, the gauge freedom is constrained. Specifically, the input scaling matrix $E$ and output scaling matrix $D$ for the block must be coupled to maintain the identity skip connection. In practice, this simplifies the canonicalization process by ``locking'' the gauge across residual blocks.

\subsection{Stateful Optimizers (Momentum/Adam)}

Standard Momentum or Adam \cite{kingma2015adam,sutskever2013momentum} maintains state variables (velocity $V$, squared gradients $M$) that are linear or quadratic in the gradient units. To maintain diagonal-scale consistency, these states must be tracked in canonical coordinates. For Momentum with coefficient $\mu$:
\begin{enumerate}
     \item Map velocity to canonical: $V' = D^{-1} V E^{-1}$.
     \item Update canonical velocity: $(V')^+ = \mu V' + \nabla_{\Wprime} \Loss$.
     \item Map back for weight update: $W^+ = W - \eta D (V')^+ E$.
\end{enumerate}

\section{Discussion: Normalization by Geometry}

\subsection{The Role of Normalization Layers}

As previously discussed, Batch Normalization (BN) and Layer Normalization (LN) are currently ubiquitous in deep learning. They function by actively manipulating the statistics of intermediate activations $h$ during the forward pass ($h \to \frac{h-\mu}{\sigma}$). Essentially, BN is intended to operate as a dynamic data-dependent gauge-fixing mechanism.
However, BN introduces significant downsides:
\begin{itemize}
     \item {\bf Batch Dependence}: It couples training examples, complicating small-batch or non-i.i.d.\ learning.
     
     \item {\bf Train/Test Mismatch}: The discrepancy between batch statistics (training) and running statistics 
               (inference) is a frequent source of bugs.
               
     \item {\bf Symmetry Breaking:} Normalization layers do not satisfy positive homogeneity 
               ($BN(\alpha x) \neq \alpha BN(x)$), breaking the clean algebraic structure of ReLU networks.
\end{itemize}
The UC-GSD framework proposes that the benefits of BN (i.e., stabilization of scale) can be achieved entirely through the optimization geometry. This suggests a path to {\em Normalization-Free Networks} \cite{brock2021nfnets} that rely on mathematical symmetry rather than patchwork architectural scaffolding.

\section{Discussion}

This work introduced the Unit-Consistent Adjoint as the geometrically-correct backward operator for networks with homogeneous nonlinearities. By acknowledging that diagonal rescalings are non-physical degrees of freedom, a preconditioned update rule was developed that is invariant to these symmetries. This formulation replaces heuristic ``normalization by architecture'' approaches with a rigorous ``normalization by geometry,'' which could potentially lead to deep learning models that are simpler and more robust. (See appendices for additional details and discussion.)


\newpage\appendix  

\section{Diagonal-Scale Equivariance via a Unit-Consistent Adjoint}
\label{sec:intro}

Deep networks built from linear maps and positively homogeneous nonlinearities (e.g., ReLU) possess a large {\em gauge freedom}: many distinct parameterizations implement exactly the same input-output function.  For example, in a bias-free ReLU multilayer perceptron:
\begin{equation*}
     h_0=x,~ z_\ell=W_\ell h_{\ell-1},~ h_\ell=\sigma(z_\ell),~ \sigma=\mathrm{ReLU},
\end{equation*}
the identity $\sigma(\alpha u)=\alpha\sigma(u)$ for all $\alpha>0$ implies a node-wise diagonal rescaling symmetry.  If $S_\ell\succ 0$ are diagonal matrices with $S_0=I$ and $S_L=I$, then the gauge transform:
\begin{equation*}
     \widetilde W_\ell ~=~ S_\ell W_\ell S_{\ell-1}^{-1} ~~ (\ell=1,\dots,L)
\end{equation*}
leaves the realized network function unchanged while rescaling hidden coordinates as $\widetilde h_\ell=S_\ell h_\ell$.
This freedom is benign at the level of representation but harmful for optimization: standard backpropagation and gradient descent are {\em not} equivariant to this symmetry and therefore can behave very differently on gauge-equivalent parameterizations.

\subsection{Why the transpose is the bottleneck}

For a linear map $y=W x$, classical backpropagation propagates errors via the Euclidean adjoint $W^{\top}$:
\begin{equation*}
     g_x ~=~ W^{\top} g_y,
\end{equation*}
where $g_y=\nabla_y\mathcal{L}$ and $g_x=\nabla_x\mathcal{L}$.
However, under a diagonal rescaling of coordinates:
\begin{eqnarray*}
      x &\mapsto& \widetilde x = E x,\\
      y &\mapsto& \widetilde y = D y,\\
      W &\mapsto& \widetilde W = DWE^{-1},
\end{eqnarray*}
for diagonal matrices $D,E\succ 0$. The transpose transforms as:
\begin{equation*}
     \widetilde W^{\top} ~=~ (DWE^{-1})^{\top} ~=~ E^{-1} W^{\top} D,
\end{equation*}
which {\em mixes} the arbitrary diagonal scales into the backward operator.  This scale mixing is precisely what makes the backpropagated signal, and consequently the gradient field in parameter space, sensitive to the choice of gauge (or to the choice of measurement units in a unit-analysis interpretation).

\subsection{Core Mechanism: The Unit-Consistent Adjoint}

The main contribution of this note is to replace the Euclidean transpose in backpropagation by a {\em unit-consistent adjoint} that is invariant (or, more precisely, consistent) under positive diagonal scalings.

Let $W\in\mathbb{R}^{m\times n}$ be a linear operator.  Define a canonical diagonal decomposition:
\begin{eqnarray*}
     W  &=& D W' E,~ D\succ 0,\ E\succ 0\ \text{diagonal}, \\
     W' &=& \mathcal{C}(W),
\end{eqnarray*}
where $\mathcal{C}(\cdot)$ is a {\em canonical} scaling map (e.g., RZ \cite{U2015,U2018}) that removes arbitrary positive row/column scalings so that $W'$ is a unique representative of the diagonal-scaling equivalence class of $W$.
(Informally: $\mathcal{C}(W)$ is the ``gauge-fixed'' form of $W$.)

\paragraph{Definition (unit-consistent adjoint).}
Given $W=D W' E$ as above, define:
\begin{equation*}
     W^{\ast_{\mathrm{UC}}} ~:=~ E^{-1} W^{\top} D^{-1}.
\end{equation*}
     Because $W^{\top} ~=~ E W'^{\top} D$, this construction simplifies to:
\begin{equation*}
     W^{\ast_{\mathrm{UC}}} ~=~ W'^{\top},
\end{equation*}
i.e., {\em take the transpose in canonical coordinates}.  In particular, if $\widetilde W$ is obtained from $W$ by arbitrary positive diagonal scalings, then $\mathcal{C}(\widetilde W)=\mathcal{C}(W)$, so:
\begin{eqnarray*}
     \widetilde W^{\ast_{\mathrm{UC}}} &=& \mathcal{C}(\widetilde W)^{\top} \\
                &=& \mathcal{C}(W)^{\top} \\
                &=& W^{\ast_{\mathrm{UC}}}. \\
\end{eqnarray*}
Thus the UC adjoint eliminates the backward-pass dependence on arbitrary diagonal gauge factors.  Conceptually, this operator is the {\em appropriate adjoint} once one acknowledges that, in homogeneous networks, diagonal rescalings of intermediate coordinates are not physically meaningful degrees of freedom and therefore should not influence the learning dynamics.

\subsection{From UC adjoint to diagonal-scale equivariant learning}
Replacing $W^{\top}$ by $W^{\ast_{\mathrm{UC}}}$ gives a diagonal-scale consistent error propagation rule for each linear map:
\begin{eqnarray*}
     g_x &=& W^{\ast_{\mathrm{UC}}} \\
     g_y &=& \mathcal{C}(W)^{\top} g_y.
\end{eqnarray*}
However, full diagonal-scale equivariance requires not only consistent backpropagated signals but also an update rule that respects the gauge action on parameters.  The UC adjoint provides the missing ingredient needed to construct such updates systematically.

The key idea is to perform optimization in the same canonical coordinates that define the UC adjoint.  With $W'=D^{-1}WE^{-1}$, the chain rule gives:
\begin{equation*}
     \nabla_{W'}\mathcal{L} ~=~ D (\nabla_W\mathcal{L}) E.
\end{equation*}
A standard gradient step in $W'$:
\begin{equation*}
     W'^+ ~=~ W' - \eta \nabla_{W'}\mathcal{L},
\end{equation*}
pushes forward to the explicit preconditioned update in the original parameters:
\begin{equation*}
     W^+ ~=~ D W'^+ E ~=~ W - \eta D^2 (\nabla_W\mathcal{L}) E^2,
\end{equation*}
which is equivariant under the diagonal gauge transformation $W\mapsto S_{\mathrm{out}}WS_{\mathrm{in}}^{-1}$ when $D,E$ are chosen canonically from $W$. The UC adjoint is therefore an {\em anchor} that determines the correct canonical coordinates, and canonical-coordinate descent then gives an update rule that preserves gauge-equivalence across iterations.

\subsection{Implications: replacing normalization-by-architecture with normalization-by-geometry}

Normalization layers (BatchNorm, LayerNorm, etc.) can be viewed as forward-pass mechanisms that forcibly control intermediate scales in order to stabilize optimization.  The program developed here proposes a different principle:
\begin{quote}
{\em Instead of modifying the architecture to manage scale, modify the backward operator and the optimization geometry so that learning is intrinsically insensitive to diagonal rescalings that do not change the function.}
\end{quote}
In summary, the UC approach makes stability a property of the learning rule rather than an artifact of additional layers and running statistics.

\subsection{Contributions and roadmap}

This program is organized around the UC adjoint as the central new mechanism:
\begin{enumerate}
\item We define the unit-consistent adjoint $W^{\ast_{\mathrm{UC}}}$ by transposing in a canonically scaled (gauge-fixed) coordinate system.

\item We use $W^{\ast_{\mathrm{UC}}}$ to construct backpropagation rules that are consistent with the diagonal rescaling symmetries implied by positive homogeneity.

\item We derive diagonal-scale equivariant optimization updates by performing descent in canonical coordinates, giving explicit preconditioned updates in the original parameters.

\item We extend the same canonical-coordinate principle to the remaining network operations (biases, convolutions, additions/residuals, concatenations/splits, and optimizer state), clarifying where exact equivariance is achievable and where architectural choices intentionally break the symmetry.
\end{enumerate}

In summary, the UC adjoint is the conceptual and technical cornerstone: it removes arbitrary diagonal scales from the backward operator, thereby enabling a coherent, symmetry-consistent optimization geometry whose downstream consequence is diagonal-scale equivariant learning.

\section{The UC Adjoint and GSD}

Consider a ReLU network (bias-free): 
\begin{eqnarray*}
     h_0 &=& x,\\
     z_\ell &=& W_\ell h_{\ell-1},\\
     h_\ell &=& \sigma(z_\ell),\\
     \sigma &=& \mathrm{ReLU},\\
     \ell &=& 1,\dots,~L,
\end{eqnarray*}
and its positive-homogeneity gauge group (positive diagonal re-scalings of hidden units):
\begin{equation*}
     S_0=I,~ S_L=I,~ S_\ell\in\mathbb{R}^{n_\ell\times n_\ell} ~\text{diagonal},~ S_\ell\succ 0,~
                               \ell=1,\dots,~L-1,
\end{equation*}
\begin{eqnarray*}
       \tilde W_\ell &=& S_\ell W_\ell S_{\ell-1}^{-1} \\
                            &\Longrightarrow& \\
        \tilde h_\ell &=& S_\ell h_\ell,\\
        \tilde z_\ell &=& S_\ell z_\ell, ~\text{and the network function is unchanged.}
\end{eqnarray*}

The Euclidean (Frobenius) gradient transformation under the gauge is:
\begin{eqnarray*}
     G_\ell &:=& \nabla_{W_\ell}\mathcal{L} \\
                &\Longrightarrow& \\
     \tilde G_\ell &=& \nabla_{\tilde W_\ell}\mathcal{L} \\
                          &=& S_\ell^{-1} G_\ell S_{\ell-1}.
\end{eqnarray*}

The RZ canonical scaling of each layer matrix is:
\begin{equation*}
     W_\ell ~=~ D_\ell W_\ell' E_{\ell-1}, 
\end{equation*}
with diagonal matrices $D_\ell\succ 0$, $E_{\ell-1}\succ 0$, and:
\begin{equation*}
     W_\ell' ~=~ \mathcal{RZ}(W_\ell).
\end{equation*}
Invariance of the canonical representative implies that the diagonal factors transform covariantly, i.e.:
\begin{eqnarray*}
     \tilde W_\ell &=& S_\ell W_\ell S_{\ell-1}^{-1} \\
                &\Longrightarrow& \\
     \tilde W_\ell' &=& W_\ell',\\
     \tilde D_\ell &=& S_\ell D_\ell,\\
     \tilde E_{\ell-1} &=& E_{\ell-1} S_{\ell-1}^{-1}.
\end{eqnarray*}
The UC adjoint is:
\begin{eqnarray*}
     W_\ell^{\ast_{\mathrm{UC}}} &:=& E_{\ell-1}^{-1}W_\ell^{\top}D_\ell^{-1} \\
                                                      &=& (W_\ell')^{\top}.
\end{eqnarray*}

Full gauge-equivariant updates requires gradient descent in 
canonical coordinates $W'$ and mapping back.
\begin{eqnarray*}
     W_\ell' &:=& D_\ell^{-1} W_\ell E_{\ell-1}^{-1} \\
            &\Rightarrow&\\
     \nabla_{W_\ell'}\mathcal{L} &=& D_\ell G_\ell E_{\ell-1}.
\end{eqnarray*}
\begin{eqnarray*}
     (W_\ell')^{+} &=& W_\ell' - \eta \nabla_{W_\ell'}\mathcal{L} \\
          &\Longleftrightarrow& \\
     W_\ell^{+} &=& D_\ell (W_\ell')^{+} E_{\ell-1}.
\end{eqnarray*}
This gives the explicit equivariant preconditioned update in the original parameters:
\begin{equation*}
\boxed{
W_\ell^{+} ~=~ W_\ell - \eta  D_\ell^{2}  G_\ell  E_{\ell-1}^{2},
                                          ~~ \ell=1,\dots,~ L
}
\end{equation*}
Equivariance is established as:
\begin{eqnarray*}
     \tilde U_\ell &:=& \tilde D_\ell^{2} \tilde G_\ell \tilde E_{\ell-1}^{2} \\
           &=& (S_\ell D_\ell)^2 (S_\ell^{-1}G_\ell S_{\ell-1}) (E_{\ell-1}S_{\ell-1}^{-1})^2 \\
           &=& S_\ell (D_\ell^{2} G_\ell E_{\ell-1}^{2}) S_{\ell-1}^{-1} \\
           &=& S_\ell U_\ell S_{\ell-1}^{-1},
\end{eqnarray*}
\begin{eqnarray*}
     \tilde W_\ell^{+} &=& \tilde W_\ell - \eta \tilde U_\ell \\
           &=& S_\ell W_\ell S_{\ell-1}^{-1} - \eta S_\ell U_\ell S_{\ell-1}^{-1} \\
           &=& S_\ell (W_\ell-\eta U_\ell) S_{\ell-1}^{-1}\\
           &=& S_\ell W_\ell^{+} S_{\ell-1}^{-1}.
\end{eqnarray*}
A condition stronger than equivariance can be obtained from the 
gauge-fixing/projection $\Pi(W) \rightarrow W$ after each step, where $\Pi$ maps 
each $W_\ell$ to its canonical representative $W_\ell'=\mathcal{RZ}(W_\ell)$.

\section{Gauge-equivariant extensions to common network operations}

We assume graph-wide viewpoint for all operations. Associate to every tensor node $v$ 
a (usually channel-wise) positive diagonal scale $S_v$.

\subsection*{Gauge action on tensors}
\begin{equation*}
     \tilde x_v = S_v x_v,~ S_v \succ 0 ~~ \text{(diagonal or block-diagonal by channels/groups)}.
\end{equation*}
A network is gauge-equivariant (and a family of parameterizations are gauge-equivalent)
iff every operation $f$ satisfies:
\begin{eqnarray*}
     \tilde y &=& f(\tilde x;\tilde\theta) \\
                  &=& S_y  f(x;\theta) \\
                  &=& S_y y.
\end{eqnarray*}
This induces constraints/transform rules for parameters $\theta$ and (necessarily) for updates.

\subsection*{Linear maps (Dense / Matrix)}

\begin{itemize}
\item Forward:
     \begin{equation*}
          y ~=~ W x,~~ x\in\mathbb{R}^{n_{\mathrm{in}}},~ y\in\mathbb{R}^{n_{\mathrm{out}}}.
     \end{equation*}

\item Gauge:
     \begin{eqnarray*}
          \tilde x &=& S_{\mathrm{in}} x,\\
          \tilde y &=& S_{\mathrm{out}} y, \\
          S_{\mathrm{in}},S_{\mathrm{out}}\succ 0~ \text{diagonal}.
     \end{eqnarray*}

\item Parameter transform for forward equivariance:
     \begin{equation*}
          \tilde W ~=~ S_{\mathrm{out}} W S_{\mathrm{in}}^{-1}.
     \end{equation*}

\item Euclidean gradient transform:
     \begin{eqnarray*}
          G &:=& \nabla_W \mathcal{L} \\
              &\Longrightarrow& \\
          \tilde G &=& S_{\mathrm{out}}^{-1} G S_{\mathrm{in}}.
     \end{eqnarray*}

\item RZ canonicalization for the layer:
     \begin{eqnarray*}
          W &=& D W' E, ~~~D\succ 0,\ E\succ 0\ \text{diagonal},\\
          W' &=& \mathcal{RZ}(W).
     \end{eqnarray*}

\item Under the gauge $W \rightarrow \tilde W = S_out W S_in^{-1}$:
     \begin{eqnarray*}
          \tilde W' &=& W',\\
          \tilde D &=& S_{\mathrm{out}} D,\\
          \tilde E &=& E S_{\mathrm{in}}^{-1}.
     \end{eqnarray*}

\item UC-adjoint:
     \begin{eqnarray*}
          W^{\ast_{\mathrm{UC}}} &:=& E^{-1} W^{\top} D^{-1} \\
                                                     &=& (W')^{\top}.
     \end{eqnarray*}

\item Gauge-equivariant gradient step obtained by descending in canonical coordinates $W'$:
     \begin{eqnarray*}
          \nabla_{W'}\mathcal{L} &=& D G E, \\
          W'^+ &=& W' - \eta \nabla_{W'}\mathcal{L},\\
          W^+ &=& D W'^+ E.
     \end{eqnarray*}

\item Explicit preconditioned update in original coordinates:
     \begin{equation*}
          \boxed{
               W^+ ~=~  W - \eta  D^{2} G E^{2}.
           }
     \end{equation*}
     (This is the same rule previously given. It is the canonical pattern to reuse.)
\end{itemize}

\subsection{Bias addition (Dense / Conv bias)}

\begin{equation*}
     z ~=~ W x + b,\qquad b\in\mathbb{R}^{n_{\mathrm{out}}}.
\end{equation*}
\begin{itemize}
\item Forward equivariance requires $b$ to scale like the output:
     \begin{equation*}
          \tilde b ~=~ S_{\mathrm{out}} b.
     \end{equation*}

\item Bias gradient and its gauge transform:
     \begin{eqnarray*}
          g_b &:=& \nabla_b \mathcal{L}\\
                 &\Longrightarrow& \\
          \tilde g_b &=& S_{\mathrm{out}}^{-1} g_b.
     \end{eqnarray*}
     
\item  Canonical bias coordinate (using the same output scaling $D$ as for $W$):
     \begin{eqnarray*}
           b &=& D b',\\
           b' &=& D^{-1} b, \\
           \nabla_{b'}\mathcal{L} &=& D g_b.
     \end{eqnarray*}
     
\item Gauge-equivariant bias update (descend in b'):
     \begin{eqnarray*}
          b'^+ &=& b' - \eta \nabla_{b'}\mathcal{L}, \\
          b^+ &=& D b'^+.
     \end{eqnarray*}
     
\item Explicit preconditioned rule:
     \begin{equation*}
          \boxed{
               b^+ ~=~ b - \eta D^{2} g_b.
          }
     \end{equation*}
\end{itemize}

\subsection{Elementwise positively-homogeneous nonlinearities}

Let $\phi$ be elementwise and positively homogeneous of degree 1:
\begin{equation*}
     \phi(\alpha u) ~=~ \alpha \phi(u) 
\end{equation*}     
for all $\alpha>0$, e.g., as holds for ReLU, leaky-ReLU, absolute value, max-pooling (channelwise scaling),
average-pooling (linear), and dropout mask (multiplicative).
\begin{eqnarray*}
     h &=& \phi(z),\\
     \phi(\alpha z) &=&\alpha\phi(z) ~~ (\alpha>0),
\end{eqnarray*}
then for any diagonal $S \succ 0$:
\begin{eqnarray*}
     \phi(S z) &=& S \phi(z), \\
                    &\Rightarrow& \\
     S_h &=& S_z.
\end{eqnarray*}
Backprop through an elementwise $\phi$ is:
\begin{equation*}
     g_z ~=~ \phi'(z)\odot g_h.
\end{equation*}
The indicator pattern is unchanged for degree-1 homogeneous $\phi$ and $S\succ 0$,
so the backprop map is gauge-consistent:
\begin{equation*}
     \tilde g_h ~=~ S^{-1} g_h ~\Longrightarrow~ \tilde g_z = S^{-1} g_z.
\end{equation*}

\subsection{Addition / Residual connections (constraints on the gauge)}

Addition node:
\begin{equation*}
     y ~=~ x_1 + x_2.
\end{equation*}
Forward equivariance requires both summands to carry the {\em same} scale:
\begin{eqnarray*}
     \tilde x_1 &=& S x_1,\\
     \tilde x_2 &=& S x_2 \\
          &\Longrightarrow& \\
     \tilde y &=& S(x_1+x_2) ~=~ S y.
\end{eqnarray*}
Therefore, an addition node imposes the constraint:
\begin{equation*}
     S_{x_1} ~=~ S_{x_2} = S_y.
\end{equation*}

\paragraph{Residual block example.}
\begin{equation*}
     h_{\ell} ~=~ h_{\ell-1} + F(h_{\ell-1};\theta).
\end{equation*}
This forces the scale on the skip path and the residual branch output to match:
\begin{equation*}
     S_{h_{\ell-1}} ~=~ S_{F\text{-out}} ~=~ S_{h_\ell},
\end{equation*}
i.e., residuals reduce the available gauge freedom and preserve equivariant updates,
but they constrain the allowable $S$'s on the computational graph.

\subsection{Concatenation / split / reshape / transpose of axes}

\begin{itemize}
\item Concatenation:
     \begin{equation*}
          y ~=~ \begin{bmatrix}x_1\\x_2\end{bmatrix}.
     \end{equation*}
     
\item Scale is block-diagonal:
     \begin{eqnarray*}
          S_y &=& \mathrm{blkdiag}(S_{x_1},S_{x_2}),\\
          \tilde y &=& S_y y.
     \end{eqnarray*}
     
\item Split is the inverse operation. Permuting axes corresponds to conjugating $S$ 
          by the permutation operator $P$:
          \begin{equation*}
               y ~=~ P x ~\Rightarrow~ S_y ~=~ P S_x P^{-1}.
          \end{equation*}
\end{itemize}
These are essentially bookkeeping rules for how the gauge acts on tensor layouts.

\subsection{Convolution (as a channelwise-scaled linear operator)}

Consider conv kernel $K$ with shape $(C_out, C_in, kH, kW)$.
\begin{itemize}
\item Channelwise gauge:
     \begin{eqnarray*}
               \tilde x &=& S_{\mathrm{in}} x,\\
                \tilde y &=& S_{\mathrm{out}} y,
      \end{eqnarray*}               
      for $S_{\mathrm{in}}\in\mathbb{R}^{C_{\mathrm{in}}\times C_{\mathrm{in}}}$,
      $S_{\mathrm{out}}\in\mathbb{R}^{C_{\mathrm{out}}\times C_{\mathrm{out}}}$ diagonal.

\item Kernel transform (broadcasted over spatial indices):
          \begin{equation*}
               \tilde K_{i,j,u,v} ~=~ (S_{\mathrm{out}})_{ii} K_{i,j,u,v} (S_{\mathrm{in}}^{-1})_{jj}.
          \end{equation*}
\end{itemize}
Treat {\em conv} as a matrix $W(K)$ after ``im2col'', then reuse the linear-map rule.
RZ scaling can be applied to the channelwise matricization of $K$, e.g. flatten ($C_out$) 
rows and ($C_in*kH*kW$) columns, or impose separate scalings on in/out channels.
The gauge-equivariant update is the same pattern with broadcasted $D^2$ and $E^2$:
\begin{equation*}
\boxed{
     K^+_{i,j,u,v} ~=~ K_{i,j,u,v} - \eta  (D^{2})_{ii}  (\nabla_K\mathcal{L})_{i,j,u,v} (E^{2})_{jj},
}
\end{equation*}
where $D$ acts on output channels and $E$ on input channels (both diagonal),
and ($\nabla_K\mathcal{L}$) is the usual conv-kernel gradient.

\subsection{Elementwise affine channel transforms (e.g., learned gains)}

\paragraph{Example.} $y = a \odot x + c$, with $a,c in R^n$ (per-channel or per-feature).
\begin{equation*}
     y ~=~ a \odot x + c.
\end{equation*}
If $x$ scales by $S$ and $y$ scales by $S$ (degree-1 equivariance), then:
\begin{itemize}
     \item ``$a$'' should be scale-invariant (dimensionless) under the gauge,
     \item ``$c$'' should scale like $y$.
\end{itemize}
\begin{equation*}
     \tilde a ~=~ a,~~ \tilde c ~=~ S c,~~ S_y=S_x=S.
\end{equation*}
The bias-like update for $c$ uses the same $D^2$ preconditioning as in (2).
For $a$, the gauge places no requirement (it is invariant), so ordinary update 
is already equivariant.

\subsection{Normalization layers are NOT degree-1 homogeneous}

LayerNorm / BatchNorm (in their usual form) may be approximately scale-invariant (training-mode, ignoring eps/running stats):
\begin{eqnarray*}
     \text{LN}(\alpha x) &\approx& \text{LN}(x), \\
     \text{BN}(\alpha x) &\approx& \text{BN}(x) ,
\end{eqnarray*}
but they {\em do not} satisfy $LN(Sx)=S LN(x)$.
In gauge language, they act like a ``gauge-fixing'' node:
\begin{equation*}
     y ~=~ \mathrm{Norm}(x)
\end{equation*}
which implies (approximately):
\begin{equation*}
     S_y ~\approx~ I ~\text{even if}~~ S_x\neq I.
\end{equation*}
The consequence of this is:
\begin{itemize}
     \item The ReLU positive-homogeneity gauge is broken or collapsed across normalization nodes.
     
     \item Exact global gauge-equivariance cannot generally hold in a network containing such nodes.
\end{itemize}
However, the linear-layer canonical updates can still be applied locally around them.

\subsection{Softmax / Sigmoid / Tanh are NOT positively homogeneous}

These {\em do not} satisfy $f(\alpha x)=\alpha f(x)$, so the multiplicative ReLU gauge
does not extend through them. (Softmax is shift-invariant: softmax($x+c1$)=softmax($x$).)
Therefore there is no nontrivial multiplicative-diagonal gauge that leaves such 
subgraphs invariant.

\subsection{Optimizers with state (Momentum / Adam): keep state in canonical coordinates}

If momentum is used, then velocity must be maintained in the canonical $W'$ coordinate:
\begin{eqnarray*}
     V' &=& D^{-1} V E^{-1},\\
     V'^+ &=& \mu V' + \nabla_{W'}\mathcal{L},\\
     W'^+ &=& W' - \eta V', \\
     W^+ &=& D W'^+ E.
\end{eqnarray*}
This preserves gauge-equivariance exactly (the same logic as the GD step),
and nalogously for Adam/RMSProp: {\em store and update all moments} $i$.

\section{Toward Eliminating Normalization ``Hacks'' via Gauge-Equivariant Training}
\label{sec:remove-norm}

A central motivation for introducing {\em gauge-equivariant} (or unit-consistent) learning rules is precisely to avoid architectural ``patches'' such as Batch Normalization (BN), Layer Normalization (LN), and their variants.  Normalization layers were originally adopted because deep networks with positively homogeneous nonlinearities (e.g., ReLU) exhibit severe {\em scale pathologies}: many distinct parameterizations represent the same function, yet standard Euclidean-gradient training behaves very differently across these equivalent parameterizations.  In practice, this manifests as instability, sensitivity to initialization, depth-dependent ill-conditioning, and a strong dependence on manual tuning.

\subsection{What normalization layers are really doing}
Normalization layers mitigate training difficulties by actively controlling intermediate scales.  For example, ignoring affine re-scaling parameters and $\varepsilon$-terms for intuition, a typical normalization step may be viewed as enforcing a constraint of the form
\begin{equation*}
     h ~\mapsto~ \widehat{h} ~\approx~ \frac{h-\mu(h)}{\sigma(h)},
\end{equation*}
so that the next layer receives an input whose typical magnitude is approximately fixed. This acts as an {\em explicit gauge-fixing mechanism}: it collapses or removes degrees of freedom corresponding to arbitrary scaling of hidden representations. However, this comes with costs::
\begin{itemize}
\item BN introduces dependence on batch statistics, coupling examples and complicating small-batch, online, or non-i.i.d.\ settings.

\item BN adds running-state (moving averages) whose training/inference mismatch can be subtle.

\item LN and related normalizations alter the symmetry structure of the network (e.g., by mixing coordinates through shared statistics), which can interfere with a clean diagonal rescaling interpretation.

\item All normalization layers add architectural and implementation complexity, plus additional hyperparameters and failure modes.
\end{itemize}

\subsection{Gauge-equivariant learning as a replacement principle}

If the {\em root cause} of training instability is the mismatch between (1) the functional invariances implied by positive homogeneity and (2) the {\em non-invariance} of standard gradient descent to these reparameterizations, then it is natural to attempt a ``first-principles'' fix:
\begin{quote}
{\em Instead of forcing activations to have convenient scale by inserting normalization layers, adjust the {\em learning rule} so that it is equivariant to the network's intrinsic scaling symmetries.}
\end{quote}

Concretely, consider a linear map (dense or convolutional) written abstractly as:
\begin{equation*}
     y ~=~ W x,
\end{equation*}
with a diagonal scaling gauge:
\begin{eqnarray*}
     x &\mapsto& S_{\mathrm{in}} x,\\
     y &\mapsto& S_{\mathrm{out}} y,\\
     W &\mapsto& \widetilde{W} \\
         &=& S_{\mathrm{out}} W S_{\mathrm{in}}^{-1},
\end{eqnarray*}
which preserves the realized function in positively homogeneous networks when combined appropriately across layers.
A gauge-equivariant (canonical) learning rule is obtained by descending in a {\em canonical coordinate} $W'$ defined through a positive diagonal decomposition:
\begin{eqnarray*}
     W  &=& D W' E, ~~ D\succ 0,~ E\succ 0 ~\text{diagonal},\\
     W' &=& \text{(canonical representative)}.
\end{eqnarray*}
Gradient descent in $W'$ induces the preconditioned update in the original parameters:
\begin{equation*}
\boxed{
     W^{+} ~=~ W - \eta  D^{2} (\nabla_W \mathcal{L}) E^{2},
}
\end{equation*}
and similarly for biases: 
\begin{equation*}
     b^{+} ~=~ b - \eta D^{2}(\nabla_b \mathcal{L}).
\end{equation*}
This update is {\em equivariant} under the diagonal rescaling gauge: if two parameter sets represent the same function via a gauge transform, then after one step of this update they remain gauge-related and thus represent the same function to first order in step size.

\paragraph{Interpretation.}
In effect, the learning rule ensures that training dynamics are not affected by arbitrary parameterization choices.  From this perspective, ``normalization layers'' attempt to solve a {\em conditioning} and {\em gauge} problem in the forward pass, whereas gauge-consistent learning entirely resolves the problem in the {\em optimization geometry}.

\subsection{Conclusion: when gauge-equivariant learning can eliminate normalization}

Under the hypothesis that normalization's primary benefit is to control scale-induced ill-conditioning (rather than to provide some fundamentally different representational capability), a successful gauge-equivariant learning rule would imply:
\begin{enumerate}
\item {\bf Normalization becomes optional:} stable training can be achieved without BN/LN by correcting the 
          optimization geometry instead of altering the network architecture.

\item {\bf Architecture simplification:} networks can be built from simpler primitives (linear maps, additions, 
          homogeneous nonlinearities) without inserting normalization layers as ``stability scaffolding.''

\item {\bf Batch-size robustness:} because the mechanism does not depend on batch statistics, performance 
          should degrade less when batches are small or data are streaming/online.

\item {\bf Parameterization robustness:} training should be less sensitive to rescalings of hidden units, to 
          initialization magnitudes, and to depth-induced scale drift.
\end{enumerate}
Although BN/LN may be regarded as non-rigorous methods intended to impose scale control, this does
not necessarily mean they do not provide {\em accidental} benefits in the form of, e.g., implicit noise
injection. The message of this note is that if such benefits are unknowingly provided, then rigorous
alternative solutions should be derived to provide them.

\end{document}